\pdfoutput=1
\PassOptionsToPackage{dvipsnames}{xcolor}
\documentclass[11pt]{article}

\usepackage[final]{acl}

\usepackage{times}
\usepackage{latexsym}
\usepackage{booktabs}
\usepackage[T1]{fontenc}

\usepackage[utf8]{inputenc}

\usepackage{microtype}
\usepackage{enumitem}
\setlist{nolistsep}

\usepackage{latexsym}
\newcommand{\cmark}{\textcolor{teal}{\ding{51}}}%
\newcommand{\xmark}{\textcolor{purple}{\ding{55}}}%
\usepackage{xspace}

\usepackage{tabularx}
\usepackage{booktabs}
\usepackage{geometry}
\newcommand{\Q}{\textbf{\color{NavyBlue}Q: }}
\newcommand{\A}{\textbf{\color{ForestGreen}A: }}

\usepackage{inconsolata}

\usepackage{graphicx}

\usepackage{adjustbox}
\usepackage{tcolorbox}
\usepackage{float}

\newcommand{\ITI}[1]{\textcolor{Red}{\textbf{#1}}}
  \definecolor{smallcolor}{HTML}{D8BFD8}
  \definecolor{mediumcolor}{HTML}{9370DB}
  \definecolor{largecolor}{HTML}{4B0082}
  \definecolor{latamcolor}{HTML}{300053}
\usepackage{latexsym}
\usepackage{booktabs}
\usepackage{graphicx}
\usepackage{amsmath}
\usepackage{pifont}
\usepackage{url}
\usepackage{hyperref}

\usepackage{algorithm}
\usepackage{algorithmic}

\usepackage{comment}

\usepackage{todonotes}  
\setuptodonotes{size=tiny,color=pink!29}

\title{Leveraging Wikidata for Geographically Informed Sociocultural Bias Dataset Creation: Application to Latin America
} 

\author{
  \textbf{Yannis Karmim*\textsuperscript{1,2,4}},
  \textbf{Renato Pino*\textsuperscript{2}},
  \textbf{Hernan Contreras*\textsuperscript{3}},
  \textbf{Hernan Lira\textsuperscript{4}},
  \textbf{Sebastian Cifuentes\textsuperscript{5}},\\
  \textbf{Simon Escoffier\textsuperscript{6}},
  \textbf{Luis Martí\textsuperscript{4}},
  \textbf{Djamé Seddah\textsuperscript{1}},
  \textbf{Valentin Barriere\textsuperscript{2,5}}\\
  \\
  \textsuperscript{1}ALMAnaCH team, Inria Paris Center;
  \textsuperscript{2}Dept. of Computer Science, Universidad de Chile;\\
  \textsuperscript{3}Institute of International Studies, Universidad de Chile;\\
  \textsuperscript{4}Inria Chile Research Center;
  \textsuperscript{5}Centro Nacional de Inteligencia Artificial;\\
  \textsuperscript{6}School of Social Work, Pontificia Universidad Católica de Chile.\\
  \small{
  * shared first authorship; 
     \textbf{Correspondence:} \url{vbarriere@dcc.uchile.cl}
  }
}
\newcommand{\ourname}{LatamQA}
\newcommand{\ourthing}{\ourname\xspace} 
\newcommand{\oursource}{\url{https://github.com/Inria-Chile/\ourname}}
\newcommand{\ourpara}[1]{\paragraph{#1}}

\begin{document}
\maketitle

\begin{abstract}
Large Language Models (LLMs) exhibit inequalities with respect to various cultural contexts. Most prominent open-weights models are trained on Global North data and show prejudicial behavior towards other cultures. Moreover, there is a notable lack of resources to detect biases in non-English languages, especially from Latin America (Latam), a continent containing various cultures, even though they share a common cultural ground.
We propose to leverage the content of Wikipedia, the structure of the Wikidata knowledge graph, and expert knowledge from social science in order to create a dataset of question/answer (Q/As) pairs, based on the different popular and social cultures of various Latin American countries. 
We create the \ourthing database of over 26k questions and associated answers extracted from 26k Wikipedia articles, and transformed into multiple-choice questions (MCQ) in Spanish and Portuguese, in turn translated to English. 
We use this MCQ to quantify the degree of knowledge of various LLMs and find out (i) a discrepancy in performances between the Latam countries, ones being easier than others for the majority of the models, (ii) that the models perform better in their original language, and (iii) that Iberian Spanish culture is better known than Latam one.\footnote{Code and datasets available at \oursource.}  
\end{abstract}

\section{Introduction and Related Work}

Disciplinary standards for ``valid'' knowledge have been concentrated in Western Europe and North America \cite{demeter2020academic}. 
%
Biases in AI systems, particularly in NLP, often originate from training data \cite{Wiegand2019}, annotation practices \cite{Sap2022}, and annotation guidelines \cite{Parmar2023}. These biases may take moral \cite{Hammerl2022}, social \cite{Sap2020}, class-based \cite{Curry2024}, or political forms \cite{Feng2023}. Although social biases can be explicitly annotated for detection and analysis \cite{Sahoo2023}, annotation is costly and highly dependent on linguistic and cultural context \cite{Fort2024,barriere-cifuentes-2024-study,barriere-cifuentes-2024-text-classifiers}, and model's biases can be sensitive to simple context changes \cite{quiroga-etal-2025-adapting}. Recent studies emphasize the importance of localizing dataset construction and warn against outsourcing bias-related annotation for non-English languages to actors in the Global North \cite{Hada2024}. Such practices risk overlooking culturally specific meanings and social dynamics, as many existing datasets inadequately represent the cultural contexts of the Global South. \cite{Santy2023}.

 \begin{figure}[t]
      \centering
      \includegraphics[width=0.8\linewidth]{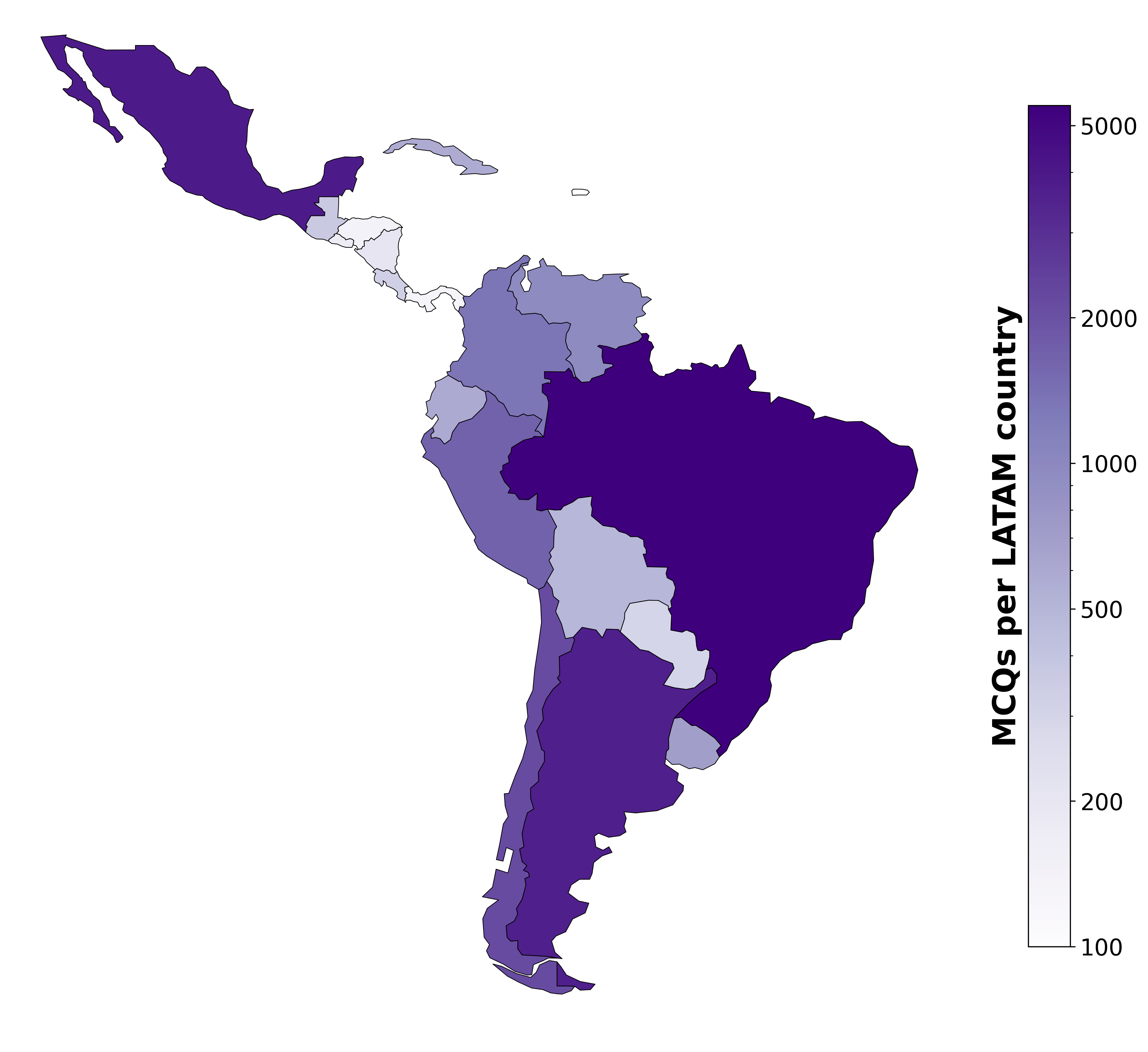}
      \vspace*{-.47cm}
      \caption{Geographic distribution of \ourthing cultural MCQs across Latin America composed of 23k Q/As.}
      \label{fig:latam_distribution}
  \end{figure}

Cultural biases in LLMs are particularly underexplored for Latin America (Latam). Although many countries share Spanish or Portuguese as official languages, they differ substantially in historical, social, and cultural terms, making the region well suited for evaluating fine-grained cultural knowledge. However, current geo-cultural datasets either group countries coarsely \cite{Czarnowska2021,Li2024k}, cover few Latam countries \cite{Myung2024}, merge heterogeneous regions \cite{Adilazuarda2024}, or are limited to English \cite{Feng2025}. In Global-MMLU, only 1.6\% of culturally relevant content concerns Latam \citep{Singh2024a}. Table~\ref{tab:latam_ds} summarizes existing cultural Q/As datasets with Latam content.

\begin{table}[tb]
    \centering
    \resizebox{\linewidth}{!}{
    \begin{tabular}{@{}l@{}c@{\,\,}c@{\,\,}c@{}}
        \toprule
        \textbf{Datasets} & \textbf{\# (k)} & \textbf{Region} & \textbf{SP/PT}  \\
        \midrule
        BLEnD \cite{Myung2024} & 2  & Mixed & \cmark \\
        CulFIT \cite{Feng2025} & .08  & Latam & \cmark \\
        CANDLE \cite{Nguyen2023a} & 2  &  Latam  & \cmark \\
        CultureBank \cite{Shi2024c} & 3.3  & Latam & \xmark \\
        CultureAtlas \cite{Fung2024} & 1.8 &  Mixed  & \xmark \\
        GLOBAL-MMLU \cite{Singh2024a} & .01 & Latam & \cmark \\
        \midrule
        \ourthing (ours) & 26 & Latam & \cmark \\
        \bottomrule
    \end{tabular}
    }
    \caption{Latam-related questions in cultural Q/As datasets in terms of number of entries, region coverage, and if in Spanish and/or Portuguese.} 
    \label{tab:latam_ds}
\end{table}

\ourpara{Cultural Benchmark Creation} Building robust datasets that capture region-specific cultural knowledge is essential for evaluating biases in LLMs \cite{Hershcovich2022a,Liu2024c,Pawar2024}. Manual annotation efforts such as BLEnD \cite{Myung2024} and CulturalBench \cite{Chiu2024} offer high-quality data but are inherently limited in scale. In contrast, automated approaches construct cultural benchmarks by extracting data from large corpora.  For example, \citet{Nguyen2023a} rely on the C4 corpus, which has been shown to contain substantial noise \cite{Fung2024}. Other methods draw on social media platforms, such as CultureBank \cite{Shi2024c}, which extracts from TikTok and Reddit or use curated web corpora, such as CRAFT \cite{Wang2024k}, which retrieves documents from SlimPajama using keywords and LLM-generated Q/As. Wikipedia offers a middle ground with clean, curated content. CultureAtlas \cite{Fung2024} extracts from Wikipedia using hyperlink expansion with NLI filtering; \citet{Li2024k} use Wikipedia categories; \citet{Zhao2025} leverage Wikidata paths. Unlike these approaches, we exploit Wikipedia's pre-existing category ontology combined with sociologist-guided validation and LLM-based Q/A generation.

\ourpara{Language and Geographic Analysis} The relationship between prompting language and cultural knowledge retrieval remains open. XNationQA \cite{Tanwar2025} finds prompting language significantly impacts performance, while \citet{Ying2025} show models perform better in native languages---though findings conflict across studies \cite{Zhao2025}. Following prior work on common ground \cite{Adilazuarda2024,Hershcovich2022a}, we focus on local facts at a fine-grained geographic level, particularly popular culture and sociocultural references that support shared understanding \cite{adams2004cultural}. Given the close relationship between language and culture \cite{Hershcovich2022a}, we evaluate models using each country's native language.

\ourpara{Our Approach} Prior datasets primarily focus on cultural norms or social practices. We target culturally grounded factual knowledge: shared references that define collective identity. We leverage Wikimedia resources together with social science expertise to construct a large-scale dataset for Latam, combining: \textit{(i)} Portuguese and Spanish alongside English to investigate language effects; \textit{(ii)} comparison between Latam and Spain to assess regional representation in training data; and \textit{(iii)} fine-grained analysis across Latam countries and cultural elements. 
Our contributions are:
\begin{itemize}
\item a scalable methodology for creating geographically informed sociocultural Q/A datasets using Wikipedia categories, expert curation, and LLM-based generation,
\item the \ourthing benchmark of 23,499 multiple-choice questions covering 20 Latam countries (see Figure~\ref{fig:latam_distribution}) in Spanish, Portuguese, and English, and
\item an empirical analysis of performance variation across countries, prompting languages, and between Latam and Iberian Spanish cultural knowledge.
\end{itemize}

\begin{figure*}[tb] \vspace*{-.2cm}
    \centering 
    \includegraphics[width=.9\linewidth]{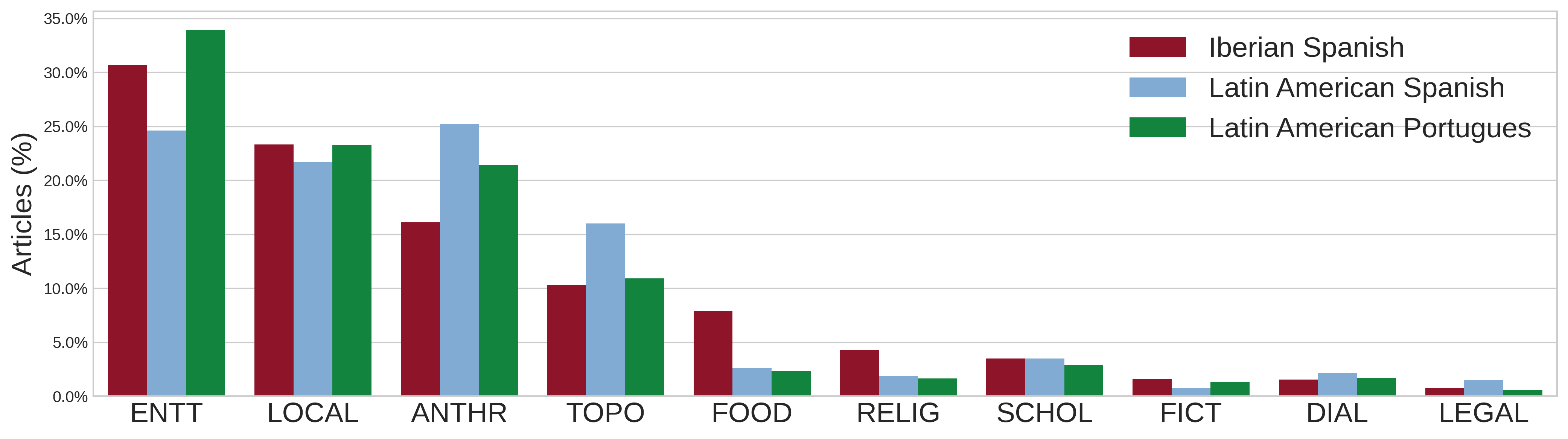} \vspace*{-.3cm}
    \caption{Distribution of the ratio of articles per cultural element per Language or Region in \ourthing. Cultural elements are: Anthroponyms (ANTHR), Forms of entertainment (ENTT), Local Institution (LOCAL), Toponyms (TOPO), Dialect (DIAL), Food and Drink (FOOD), Legal System (LEGAL), Scholastic reference (SCHOL), Religious celebration (RELIG), Fictional character (FICT).} 
    \label{fig:cultural_distribution}
\end{figure*}

\section{Benchmark Creation} 

%

\subsection{Raw Wikipedia Data} 

We apply two sociology-based filters to enhance the dataset pertinence: one at the category level and one at the article level.

\paragraph{Collection} Our data collection method relies on the Wikipedia categories' ontology.  
As each category contains articles and subcategories, it is possible to scrap the content in a recursive way, and, therefore, obtain a structured list of articles with associated metadata. We start from a mother category containing cultural information about a Region of Interest (RoI) such as ``\emph{Cultura de Chile},'' ``\emph{Cultura de Peru},'' or other RoI, and recursively collect the links of the Wikipedia articles and subcategories (see Algorithm \ref{alg:scraping} in Appendix \ref{app:scraping}).

A manual validation of the main subcategories\footnote{up to three layers of depth inside the ontology} from a sociologist helps removing the categories that are not relevant for a RoI, such as ``\emph{Idioma Español}'' which contains everything related to Spanish language in general, or ``\emph{Alumnados de \texttt{[ENT]}}'' which contains all the people that went to the school \texttt{[ENT]}. 
%
This allowed us to obtain 154k articles. 
Metadata from Wikimedia was used to filter out the documents not relevant to the specific country.  

\paragraph{Curation}
%
%
Not all articles contained within the remaining subcategories are equally relevant to cultural analysis.
To address this, we apply a second filter at the article level by manually annotating each article according to its socio-cultural relevance. We define three classes: positive, descriptive, and negative. 
%
%
%
The negative class includes articles that do not address any cultural elements defined in the taxonomy proposed by \citet{Espindola2006}. 
Articles that are culturally relevant are assigned to either the positive or descriptive classes.
The descriptive class is used for articles that primarily contain technical or enumerative information details\footnote{List of the football teams, statistics, transfer dates of a player \emph{vs.} political history, details on the anthem, rivalries with opponents of a club.} with limited interpretive value, such as lists of songs from a specific music album. 
%
500 articles were manually tagged and used to fine-tune a pre-trained multilingual Longformer \cite{Beltagy2020-bh}, reaching a precision of 87.5\% with respect to the positive class, and 100\% when merging the positive and descriptive classes.
%
Details are available in Appendix \ref{app:automatic_filtering}. 


\paragraph{Cultural Elements Distribution} Wikipedia articles are generally associated with metadata specifying their entity type within an ontology \cite{vrandevcic2012wikidata}. We obtained 2,169 distinct entities across the entire dataset. Using an LLM (Qwen3-Max), 
we tagged the entities in a zero-shot in-context-learning way with rapid manual verification, mapping each Wikidata entity type to one of our predefined cultural elements. The resulting distribution of articles across cultural elements is presented in Figure \ref{fig:cultural_distribution}.

\subsection{Questions and Answers Generation} 



\begin{table*}[tb]
\centering
\tiny
\setlength{\tabcolsep}{4pt}
\begin{tabular}{@{}>{\bfseries}lp{0.92\linewidth}@{}}
\toprule
\textbf{Category} & \textbf{Questions and Answers} \\
\midrule

FICT & 
\Q What role did the \textit{Bacab} play in Maya beekeeping? 
\A They were the primary protectors of bees and founders of apiculture. \\


& 
\Q What is \textit{glíglico}, and in which literary work does it appear? 
\A \textit{Glíglico} is a fictional language created by Julio Cortázar and appears in his novel \textit{Rayuela}.
\\
\midrule

FOOD & 
\Q In which Mexican state is the \textit{memela} considered a traditional dish? 
\A The \textit{memela} is a traditional dish from the state of Puebla. \\


& 
\Q What cultural origins does Lima’s \textit{mazamorra morada} have in Peru? 
\A Lima’s \textit{mazamorra morada} has Afro-Peruvian roots and is part of Peru’s culinary identity. \\

\midrule

DIAL & 
\Q What old expression used in Mexico City means “it seems to me” and remains in everyday speech? 
\A The expression “\textit{se me hace}” is used in Mexico City to mean “it seems to me.” \\


& 
\Q In Chile, which social group is the term “\textit{flaite}” disparagingly directed toward? 
\A It refers to lower-class individuals who are socially inadapted and aggressive, and also to any vulgar behavior regardless of social origin. \\

\midrule

%
RELI & 
\Q What is the role of the \textit{machi} in Mapuche culture?
\A The \textit{machi} is a Mapuche medicine central figure, healing physical and spiritual ailments, with religious and social functions \\


& 
\Q On which day of the month is the tradition of eating \textit{ñoquis} observed in Argentina, Uruguay, Brazil, and Paraguay? \A It is celebrated on the 29th of each month. \\

\midrule

ENTTT & 
\Q According to the article, what is the origin of \textit{cumbia}? 
\A \textit{cumbia} results from a mixture of Indigenous and African influences. \\


& 
\Q In which Mexican state is the novel \textit{Falsa liebre} set? 
\A The novel is set in Veracruz, showing the most marginalized and violent side of the region. \\
\bottomrule
\end{tabular}
\caption{Example of questions and answers from different cultural elements.} \label{tab:questions_examples}
\end{table*}

We leverage the filtered Wikipedia articles database to generated article-grounded questions and associated answers. Several prompting strategy were evaluated, with respect to a topic-dependant definition of culture that would apply the most to extract interesting knowledge from the Wikipedia data. \texttt{gpt-oss-120b} was used during this phase. Examples of questions and answers are available in Table \ref{tab:questions_examples}. 


\paragraph{General Prompts} We compared several prompts to generate questions grounded with various definitions of cultures, based on: anthropology, general cultural exploration, psychological and symbolic significance, sociology, or on an integrative cultural definition. 
To select the definition leading to the most pertinent questions, they were manually validated by a sociologist with respect to their simplicity, quality and sociological pertinence. 
Details in Appendix \ref{app:cultural_prompts}. 

\paragraph{Questions Generation and Validation}
Once the culture definition fixed, we designed a prompt for article-grounded extraction of questions and associated answers in a structured way. 
We quantitatively validated the socio-cultural pertinence of the questions using a three-dimension notation based on \cite{geertz1973chapter,hudson2009symbolic,paezrovira2007memoria-fixed} that : (i) symbolic, (ii) social practices, and (iii)  social representations, memory and identity. We found that 98\% of the questions were at least relevant in two of the three dimensions. 
%
We also validated quantitatively the answers' grounding to the article on 100 examples, and found no case of hallucination. Details in Appendix \ref{app:validation_qa}.

\begin{figure}[tb]  
    \centering
    \includegraphics[width=\linewidth]{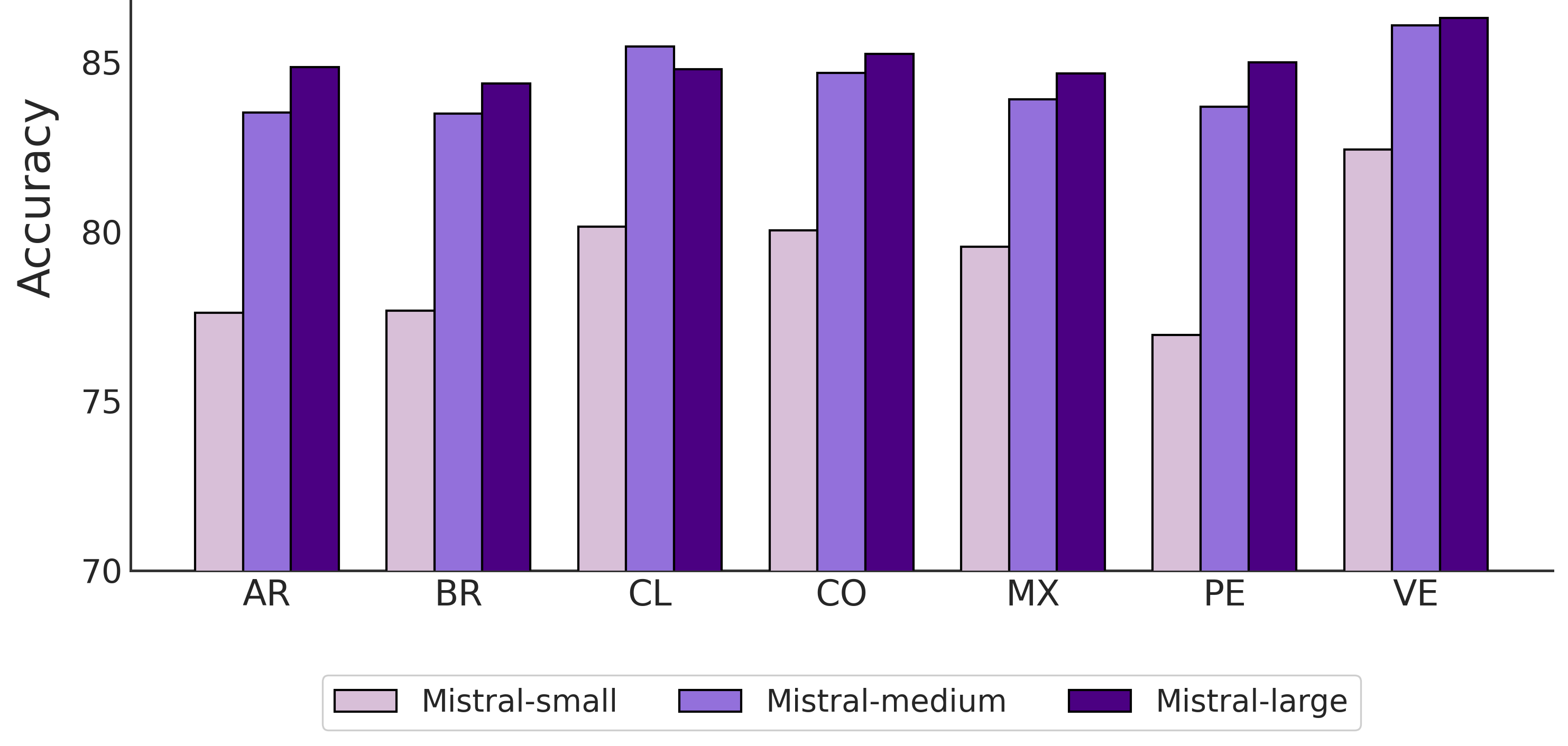}
    \vspace*{-.74cm}
    \caption{Cross-country performance of Mistral models on \ourthing. Scaling from Small to Large yields consistent improvements (+5 -- +8\% accuracy).} 
    \label{fig:perf_per_big_country}
\end{figure}

%


\paragraph{Distractors Generations}
We generated distractors as \citet{Fung2024}, 
generating alternative answers with the same LLM that generated the questions. 
Details in Appendix \ref{app:mcqa_gen}. 

\begin{table}[tb] 
    \centering
    \small
    \definecolor{smallcolor}{HTML}{C19AC1}  
    \definecolor{mediumcolor}{HTML}{9370DB}
    \definecolor{largecolor}{HTML}{4B0082}
    \setlength{\tabcolsep}{3.8pt}  
    \begin{tabular}{@{}lcccc|cc@{}}
        \toprule
        & \multicolumn{2}{c}{\textbf{Brazilian PT}} & \multicolumn{2}{c}{\textbf{Latam SP}} & \multicolumn{2}{c}{\textbf{Spain}} \\
        \cmidrule(lr){2-3} \cmidrule(lr){4-5} \cmidrule(lr){6-7}
        \textbf{Model} & PT & EN & SP & EN & SP & EN \\
        \midrule
        \multicolumn{7}{@{}l}{\textit{\textcolor{smallcolor}{\textbf{Small models}}}} \\
        Llama 3.1-8B & 65.9 & 66.2 & 69.2 & 64.5 & 76.0 & 80.5 \\
        Mistral-small& 77.0 & 74.3 & 78.5 & 76.1 & 84.3 & 81.4 \\
        \midrule
        \multicolumn{7}{@{}l}{\textit{\textcolor{mediumcolor}{\textbf{Medium models}}}} \\
        Qwen2.5-14B & 65.1 & 62.1 & 68.8 & 67.5  & 79.1 & 78.2 \\
        GPT-4.1-mini & 80.0 & 76.1 & 81.5 & 78.2 & \textbf{88.0} & 85.1 \\
        Mistral-medium & 82.6 & 81.8 & 83.9 & 80.5 & 87.1 & {85.4} \\
        \midrule
        \multicolumn{7}{@{}l}{\textit{\textcolor{largecolor}{\textbf{Large models}}}} \\
        Qwen3-430B & 70.8 & 71.4 & 75.8 & 74.0  & 83.7 & 82.4 \\
        Kimi-K2-thinking & 69.6 & 70.5 & 71.6 & 70.9 & 81.0 & 76.1 \\
        Mistral-large & \textbf{84.3} & \textbf{83.0} & \textbf{85.4 }& \textbf{81.8} & 87.6 & \textbf{86.4} \\
        \midrule
        \multicolumn{7}{@{}l}{\textit{\textcolor{latamcolor}{\textbf{Latam Model}}}} \\
        PatagonIA & 81.5 & 76.8 & 82.0 & 79.2 & 86.9 & 84.9 \\
        LatamGPT & -- & -- & -- & -- & -- & -- \\
        \bottomrule
    \end{tabular}
    \caption{Performance of various LLMs on the \ourthing benchmark (accuracy \%). We evaluate models with both native languages (PT and SP) and MT English translations.
    } \label{tab:benchmark_mcq}
\end{table}

\section{Experiments and Results} \vspace*{-.1cm}



\paragraph{Global Results and Prompting Language} 
The performances of various size models are visible in Table \ref{tab:benchmark_mcq} and in the same range of other culture-related datasets \cite{Myung2024,Ying2025}. 
All the models are performing better in their native language (ES or PT), which is consistent with past results as Spanish and Portuguese are high-resource languages \cite{Myung2024,Ying2025} but contradictory with other works \cite{Tanwar2025,Zhao2025}. We believe that might be due to the nature of the source document to create the Q/A (graph triplet) or because of heterogenous generation capabilities between languages \cite{Kabir2025}. 
We also include PatagonIA \cite{patagonia-ia} and LatamGPT \cite{LatamGPT},
\footnote{Results to be added soon using the official release.} the former is specialized in Chilean Spanish, supposedly based on a Sparse-MoE architecture, the latter is a Llama 3.1 70B that has been pre-trained again over 300 billion tokens spanning Spanish, English, and Portuguese—with a significant portion of the data sourced directly from various countries across the Latam region.

%
Despite its regional specialization, PatagonIA does not outperform general-purpose models of medium size such as Mistral-medium or GPT-4.1-mini. 


\paragraph{Model Size \emph{vs.} Performance} 
While performances are heterogenous with respect to the LLMs, it is notable that they are homogenous with respect to the size of the model. We can notice consistent improvements, with the exception of Mexico, on the biggest countries for various Mistral models in Figure \ref{fig:perf_per_big_country}.\footnote{Same phenomena observed for \texttt{Qwen2.5} and \texttt{Qwen3}.}

\paragraph{Iberian \emph{vs.} Latam Spanish} 
Using a similar process, we extracted a set of Q/As from Spain to compare the performances of the models. All the models performed better on the Iberian Spanish subset. The results are coherent with \citet{Myung2024}, where the models reached higher performance on questions from Spain than questions Mexico subset. It is interesting to note that even if Mistral models still perform very well, now the best results are obtained with \texttt{GPT-4.1}. 

\begin{figure}[tb] 
    \centering
    \includegraphics[width=\linewidth]{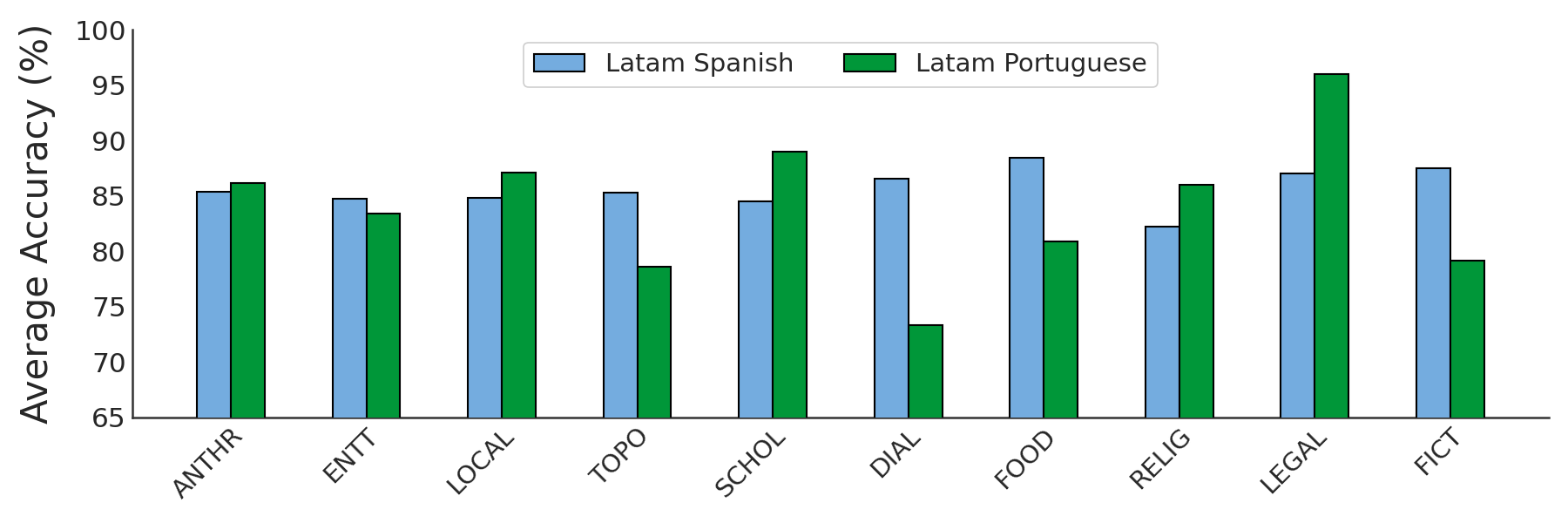}
    \caption{Performance of Mistral-large in Latam Spanish and Portuguese with respect to the different cultural elements.} \vspace*{-.3cm}
    \label{fig:perf_per_cultural_element_gap}
\end{figure}

\paragraph{Cultural Element-level Analysis} 
Leveraging Wikidata's type of entity ontology, we automatically map every article and its associated question to its cultural element in the taxonomy of \citet{Espindola2006}. 
Performance gaps (see Figure~\ref{fig:perf_per_cultural_element_gap}) between Latam Spanish and Brazilian Portuguese are higher when the cultural elements contain few examples such as ``Fictional character'' and ``Dialect''  where we observed between 75\% and 80\% of accuracy for Brazilian Portuguese.

\section{Conclusion} 

This work introduces a new sociocultural benchmark focused on Latam. The benchmark contains more than 23k distinct multiple-choice questions (MCQs), each paired with a ground-truth answer. 
We construct this large-scale, structured dataset by combining information from Wikipedia and Wikidata with domain expertise from the social sciences. First, LLM performance varies substantially across models, although it remains consistent across different scales within the Mistral family. Second, using the native language of the target culture leads to better performance for Spanish and Portuguese. Third, all evaluated LLMs perform better in Iberian Spanish than in Latin American Spanish. Fourth, an analysis at the level of cultural elements shows that performance varies depending on the type of knowledge being tested. 
Taken together, the results indicate that LLMs can operationalize epistemic injustice: they deliver systematically higher reliability for contexts already more visible in dominant information infrastructures while degrading for underrepresented ones even when language is held constant. Because LLMs are increasingly used as default knowledge interfaces, these asymmetries risk automating gatekeeping and amplifying existing global gradients in recognition and authority.

\section{Limitations}


This work represents an initial step toward estimating the cultural knowledge of large language models (LLMs) in South America. However, cultural knowledge cannot be adequately captured through simple prompt-based question answering alone \cite{Zhou2025a,Kabir2025}. Future work should therefore move beyond basic multiple-choice question answering (MCQA) benchmarks \cite{Oh2025}. Promising directions include directly involving human participants in benchmark construction \cite{Ivetta2025,Ivetta2025a} and analyzing interactional data, such as discussions in Wikipedia Talk Pages associated with the target articles.. Similarly, we are aware of the possible preference biases \cite{Wataoka2025-xk} that might be introduced by using only one LLM. 
%

\section*{Acknowledgments}

This work was partially financed with the grant U-INICIA 2024 from the Vicerrectoría de Investigación y Desarrollo (VID) number UI-011/24 “Estudios de sesgos sociales en modelos de lenguajes largos", by the Franco-Chilean Binational Center of Artificial Intelligence, ANID Strengthening R\&D capabilities Program CTI230007 Inria Chile. This work was granted access to the HPC resources of IDRIS under the allocation 2025-A0180616119 made by GENCI.
%
We thank the Patagonia IA  and LatamGPT teams for sharing the result of their models.  


\bibliography{good}

\appendix
\setcounter{figure}{0} 
\setcounter{algorithm}{0} 
\setcounter{table}{0} 
\renewcommand\thefigure{\thesection.\arabic{figure}}    
\renewcommand\thealgorithm{\thesection.\arabic{algorithm}} 
\renewcommand\thetable{\thesection.\arabic{table}}

\section{Scraping Algorithm}
\label{app:scraping}

The scraping algorithm is described in Algorithm \ref{alg:scraping}.  We collected articles from the category pages titled ``\emph{Cultura de \texttt{[Country]}}'' for the 20 countries listed in Table \ref{tab:questions_distribution} using each country’s main language. 
We set MAX\_DEPTH to 5 as empirical testing showed that greater depths reduced the relevance of the retrieved articles. For Spain, we used \verb|MAX_DEPTH = 3| due to the substantially larger initial number of articles. 

\begin{algorithm*}[tb]
\caption{Recursive Wikipedia Category Scraper}
\begin{algorithmic}[1]
\STATE \textbf{Input}: Initial Wikipedia category URL, Maximum recursion depth MAX\_DEPTH

\STATE \textbf{function} \textsc{ScrapeCategory}(categoryURL, currentDepth):
\begin{ALC@g}
    \IF {currentDepth > MAX\_DEPTH}
        \RETURN
    \ENDIF
    \STATE Fetch HTML category page to extract articles and subcategory links 

    \FOR{each article link NOT already processed}
        \STATE Fetch HTML article page content and save article data
    \ENDFOR

    \FOR{each subcategory link}
        \STATE 
        \textsc{ScrapeCategory}(subcategoryURL, currentDepth + 1)
    \ENDFOR
\end{ALC@g}

\STATE \textsc{ScrapeCategory}(initialCategoryURL, currentDepth = 0)

\end{algorithmic}
\label{alg:scraping}
\end{algorithm*}

\setcounter{figure}{0} 
\setcounter{algorithm}{0} 
\setcounter{table}{0} 
\section{Automatic Filtering} \label{app:automatic_filtering}

\subsection{Elements of Culture}

We base our filtering on the elements of culture from \citet{Espindola2006} used to filter out articles in the negative class. 

\subsubsection{Definitions}

\begin{itemize}
    \item   \textbf{TOPO} (Toponyms): a place name, a geographical name, a proper name of locality, region, or some other part of Earth’s surface or its natural or artificial feature.
    \item   \textbf{ANTHR} (Anthroponyms): ordinary and famous people’s names and nick-names and names referring to regional background which acquire identification status; it also includes animals that have been given human qualities, symbols, or political meanings in social representations.
    \item   \textbf{ENTT} (Forms of entertainment): amusement or diversion including public performances or shows, it also encompasses hospitality provided, such as dinners, parties, business lunches, etc.; it also includes artistic expressions.
    \item   \textbf{FICT} (Fictional character): a person in a novel, play, or a film who is related to fiction, works of imagination.
    \item   \textbf{LEGAL} (Legal System): rules of conduct inherent in human nature and essential to or binding upon human society.
    \item   \textbf{INST} (Local Institution): an organization that helps or serves people in a certain area - health, education, work, political, administrative, religious, artistic; it also includes national symbols.
    \item   \textbf{FOOD} (Food and Drink): any solid or liquid substance that is used by hu-man beings as a source of nourishment.
    \item   \textbf{SCHOL} (Scholastic reference): related to school or studying.
    \item   \textbf{RELIG} (Religious celebration): to do something special to mark a religious occasion.
    \item   \textbf{DIAL} (Dialect): user-related variation, which determines speaker’s status as regards social class, age, sex, education, etc.; it also includes slang.
\end{itemize}

\subsubsection{Distribution}

The distributions of the articles with respect to its cultural element relevance is shown in Figure \ref{fig:cultural_distribution}. The biggest difference lies within the ratio of articles about Food and Drink: they are more dominant in Spain than in Latam.

\subsection{Classifier} \label{app:longformer}

We fine-tuned and validated a pre-trained XLM-RoBERTa Longformer\footnote{\texttt{markussagen/xlm-roberta-longformer-base-4096}} on 500 3-class examples. When merging the descriptive and positive classes, the classifier reaches an accuracy of 97.8\%. The confusion matrix obtained from cross-validation is the following ($c_{i,j}=c_{y,\hat{y}}$): 

\begin{equation*}
\begin{bmatrix}
198 & 24 & 0 \\
43 & 98 & 19 \\
11 & 40 & 131
\end{bmatrix}
\end{equation*}

\setcounter{figure}{0} 
\setcounter{algorithm}{0} 
\setcounter{table}{0} 
\section{Per-Country Distribution}

The distribution of the dataset questions per country is shown in Table \ref{tab:questions_distribution}. 

\begin{table}[tb]
\centering
\label{tab:dataset_stats}
\small
\resizebox{\linewidth}{!}{
\begin{tabular}{@{}llr@{}}
    \toprule
    \textbf{Country/Region} & \textbf{Language} & \textbf{Count} \\
    \midrule
    Brazil (BR) & Portuguese & 6,075 \\
    México (MX) & Spanish & 4,893 \\
    Argentina (AR) & Spanish & 4,243 \\
    Chile (CL) & Spanish & 2,469 \\
    Perú (PE) & Spanish & 1,921 \\
    Colombia (CL) & Spanish & 1,752 \\
    Brazil (BR) & Spanish & 1,164 \\
    Venezuela (VE) & Spanish & 1,030 \\
    Cuba (CU) & Spanish & 674 \\
    Ecuador (EC) & Spanish & 720 \\
    Uruguay (UY) & Spanish & 991 \\
    Bolivia (BO) & Spanish & 750 \\
    Guatemala (GT) & Spanish & 743 \\
    Costa Rica (CR) & Spanish & 467 \\
    El Salvador (SV) & Spanish & 306 \\
    Nicaragua (NI) & Spanish & 436 \\
    Paraguay (PY) & Spanish & 542 \\
    Dominican Republic (RD) & Spanish & 234 \\
    Honduras (HN) & Spanish & 180 \\
    Panamá (PA) & Spanish & 218 \\
    Puerto Rico (PR) & Spanish & 193 \\
    \midrule
    \multicolumn{2}{@{}l@{}}{\textbf{Total}} & \textbf{26,213} \\
    \bottomrule
\end{tabular}
}
\caption{
Distribution of articles from the \ourthing dataset that are from a categories under the mother category \texttt{"Cultura de [Country]"}. In this case, an article can be associated with several countries, and possibly languages.} \label{tab:questions_distribution}
\end{table}

\setcounter{figure}{0} 
\setcounter{algorithm}{0} 
\setcounter{table}{0} 
\section{Questions-Generation Prompts}
\label{app:cultural_prompts}

\subsection{Domain-specific Culture Definition}

We first selected a prompt from five different prompts using different definitions of culture: an anthropological approach, general cultural exploration approach, psychological and symbolic significance approach, sociological approach and integrative cultural definition approach. 
The quality of the questions was assessed with respect to the clarity of language under ``theoretical principles of phenomenology, which studies things as they are shown in consciousness to make them comprehensible'' (\citealt{lugo2025innovacion}; page 188), which means expressing the questions in simple terms. 
%
A good question is correctly formulated, asking for something precise and not ambiguous present in the article, not using an overly complex or specific vocabulary or concepts (such as ``collective identity'' or ``communal expression'').

The general cultural exploration approach was judged the most relevant for the benchmark creation (see Figure \ref{fig:prompt-cultural-approach}). 

  \begin{figure*}[tb]
  \centering
  \small
  \begin{tcolorbox}[
      colback=smallcolor!15,
      colframe=mediumcolor!70,
      title=\textbf{General Cultural Exploration},
      fonttitle=\bfseries\small,
      coltitle=largecolor,
      colbacktitle=smallcolor!40,
      boxrule=0.5pt,
      arc=2pt,
      left=6pt,
      right=6pt,
      top=4pt,
      bottom=4pt
  ]
    \textbf{System:} From the given Wikipedia article, identify and formulate relevant sociocultural questions and answers.
    
    \vspace{0.5em}
    Questions should address topics such as cultural identity, popular symbols, collective memory, traditions, and practices specific to the local community or region described. 

    \vspace{0.5em}
    Answers must briefly summarize the sociocultural relevance of these elements
  \end{tcolorbox}
  \caption{General Cultural Exploration approach prompt.}
  \label{fig:prompt-cultural-approach}
  \end{figure*}

 \begin{figure*}[tb]
  \centering
  \small
  \begin{tcolorbox}[
      colback=smallcolor!15,
      colframe=mediumcolor!70,
      title=\textbf{Cultural QA Generation Prompt (Spanish)},
      fonttitle=\bfseries\small,
      coltitle=largecolor,
      colbacktitle=smallcolor!40,
      boxrule=0.5pt,
      arc=2pt,
      left=6pt,
      right=6pt,
      top=4pt,
      bottom=4pt
  ]
  \textbf{System:} You are an expert in cultural anthropology and educational content creation.

  Your task is to generate three culturally meaningful question--answer pairs in Spanish based only on information explicitly present in the following Wikipedia article.

  \vspace{0.5em}
  \textbf{\textcolor{largecolor}{Goal:}} Identify the most culturally significant element explicitly mentioned in the article and produce three question--answer pairs about it, focusing on:
  \begin{itemize}[noitemsep,topsep=2pt,leftmargin=*]
      \item Cultural identity
      \item Popular symbols
      \item Collective memory
      \item Traditions
      \item Practices specific to the local community
  \end{itemize}

  \vspace{0.5em}
  \textbf{\textcolor{largecolor}{Rules:}}
  \begin{itemize}[noitemsep,topsep=2pt,leftmargin=*]
      \item Use only information explicitly contained in the article.
      \item Select the most widely recognized cultural element mentioned.
      \item The question must be clear and culturally meaningful.
      \item The answer must be brief and concise, one or two sentences at most.
      \item Output must follow exactly the JSON format below.
      \item Do not invent information.
  \end{itemize}

  \vspace{0.5em}
  \textbf{\textcolor{largecolor}{Article:}} \texttt{\{content\}}

  \vspace{0.5em}
  \textbf{\textcolor{largecolor}{Output Format:}}
  \begin{verbatim}
  {"question1": "...", "answer1": "...",
   "question2": "...", "answer2": "...",
   "question3": "...", "answer3": "..."}
  \end{verbatim}
  \end{tcolorbox}
  \caption{Prompt template for generating culturally grounded question--answer pairs from Spanish Wikipedia articles.}
  \label{fig:prompt-qa}
  \end{figure*}
  \begin{figure*}
  \centering
  \small
  \begin{tcolorbox}[
      colback=smallcolor!15,
      colframe=mediumcolor!70,
      title=\textbf{MCQ Distractor Generation Prompt (Spanish)},
      fonttitle=\bfseries\small,
      coltitle=largecolor,
      colbacktitle=smallcolor!40,
      boxrule=0.5pt,
      arc=2pt,
      left=6pt,
      right=6pt,
      top=4pt,
      bottom=4pt
  ]
  \textbf{System:} You are an expert in creating difficult multiple-choice questions to evaluate reading comprehension and cultural knowledge.

  \vspace{0.5em}
  \textbf{\textcolor{largecolor}{Context:}}\\
  Article: \texttt{\{content\}} \quad Question: \texttt{\{question\}} \quad Correct answer: \texttt{\{answer\}}

  \vspace{0.5em}
  \textbf{\textcolor{largecolor}{Task:}} Generate exactly 3 incorrect answers (distractors) in Spanish that are:
  \begin{itemize}[noitemsep,topsep=2pt,leftmargin=*]
      \item Plausible and similar in length/style to the correct answer
      \item Culturally believable (could reasonably be confused with the correct answer)
      \item Difficult to eliminate without having carefully read the article
      \item Grammatically correct in Spanish
  \end{itemize}

  \vspace{0.5em}
  \textbf{\textcolor{largecolor}{Strategies to use (vary across the 3 distractors):}}
  \begin{enumerate}[noitemsep,topsep=2pt,leftmargin=*]
      \item \textit{Entity swap}: Use another cultural element, place, or name mentioned in the article
      \item \textit{Temporal/geographical confusion}: Use a date, period, or location from elsewhere in the article
      \item \textit{Partial truth}: Capture part of the cultural context but make false on the key element
      \item \textit{Cultural plausibility}: Reference a related but incorrect tradition from the same region
      \item \textit{Subtle relation inversion}: Invert origin/influence, cause/effect, or historical sequence
  \end{enumerate}

  \vspace{0.5em}
  \textbf{\textcolor{largecolor}{Constraints:}} Each distractor must fool a quick reader; avoid absurd answers and obvious negations; match the tone of the correct answer.
  \end{tcolorbox}
  \caption{Prompt template for generating challenging distractors for multiple-choice questions.}
  \label{fig:prompt-mcq-gen}
  \end{figure*}

\subsection{MCQA Generation} \label{app:mcqa_gen}

Using general cultural exploration approach, a new prompt was designed to extract the pairs of Q/As, adding specific rules to force the questions and answer to be precise, explicit, pertinent and generated in a specific format (see Figure \ref{fig:prompt-qa}). 
A good answer responds totally to the question, uses solely the content of the article without adding external facts, and does not add specific reasoning \cite{Raphael1986,Grice1975}.  

Second, following the methodology of \cite{Fung2024}, another prompt was used to generate challenging counterfactual answers for the MCQ, which we call distractors~(see Figure \ref{fig:prompt-mcq-gen}). 

\setcounter{figure}{0} 
\setcounter{algorithm}{0} 
\setcounter{table}{0} 
\section{Validation of Q/As}
\label{app:validation_qa}

We asked two experts to score 100 questions with a 5-point likert scale with respect to the symbolic, the social practices, and the social representations, memory and identity. Only two questions over the 100 obtained a score less than 5, which means that only 2\% of the questions obtained a score of 1/1/1 or 1/1/2 and were rejected. Inter-annotator-agreement was high.

\setcounter{figure}{0} 
\setcounter{algorithm}{0} 
\setcounter{table}{0} 
\section{Mistral Models Performances} \label{app:models_perf}



\begin{figure*}[tb]
    \centering
    \includegraphics[width=\linewidth]{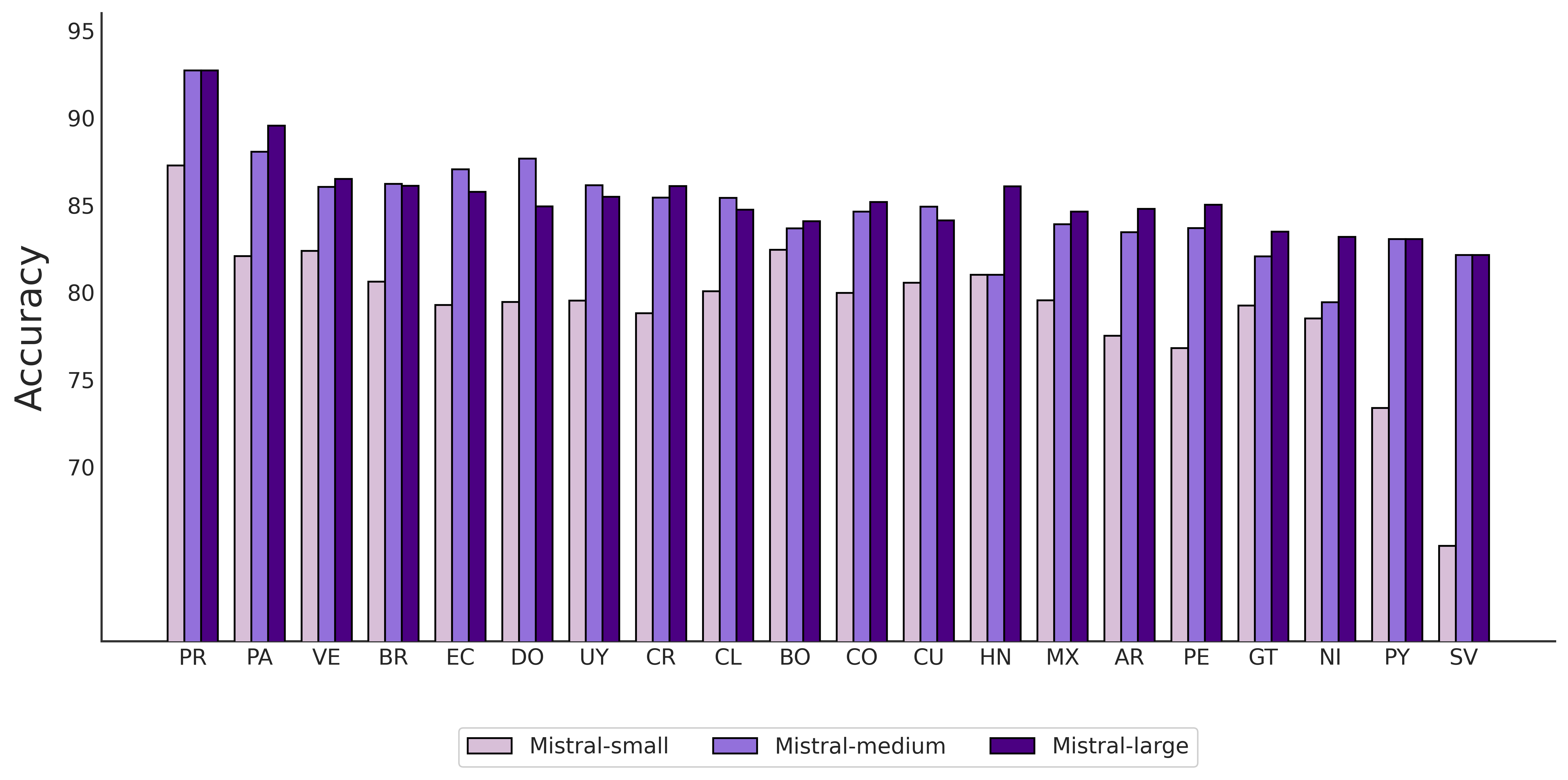}
    \caption{Cross-country performance of Mistral models on cultural knowledge evaluation. Scaling from Small to Large yields consistent improvements (+5-8\% accuracy).}
    \label{fig:perf_per_country}
\end{figure*}

\paragraph{Cross-country Performances}
The full performances of the models from the Mistral family are visible in Figure \ref{fig:perf_per_country}. It is visible that the scale consistently helps in reaching higher performances. Except for a few countries where the medium excels slightly the large model: Costa Rica, Honduras and Ecuador. 

\paragraph{Cultural Element-level Performances}

\begin{figure*}[tb]
    \centering
    \includegraphics[width=0.83\linewidth]{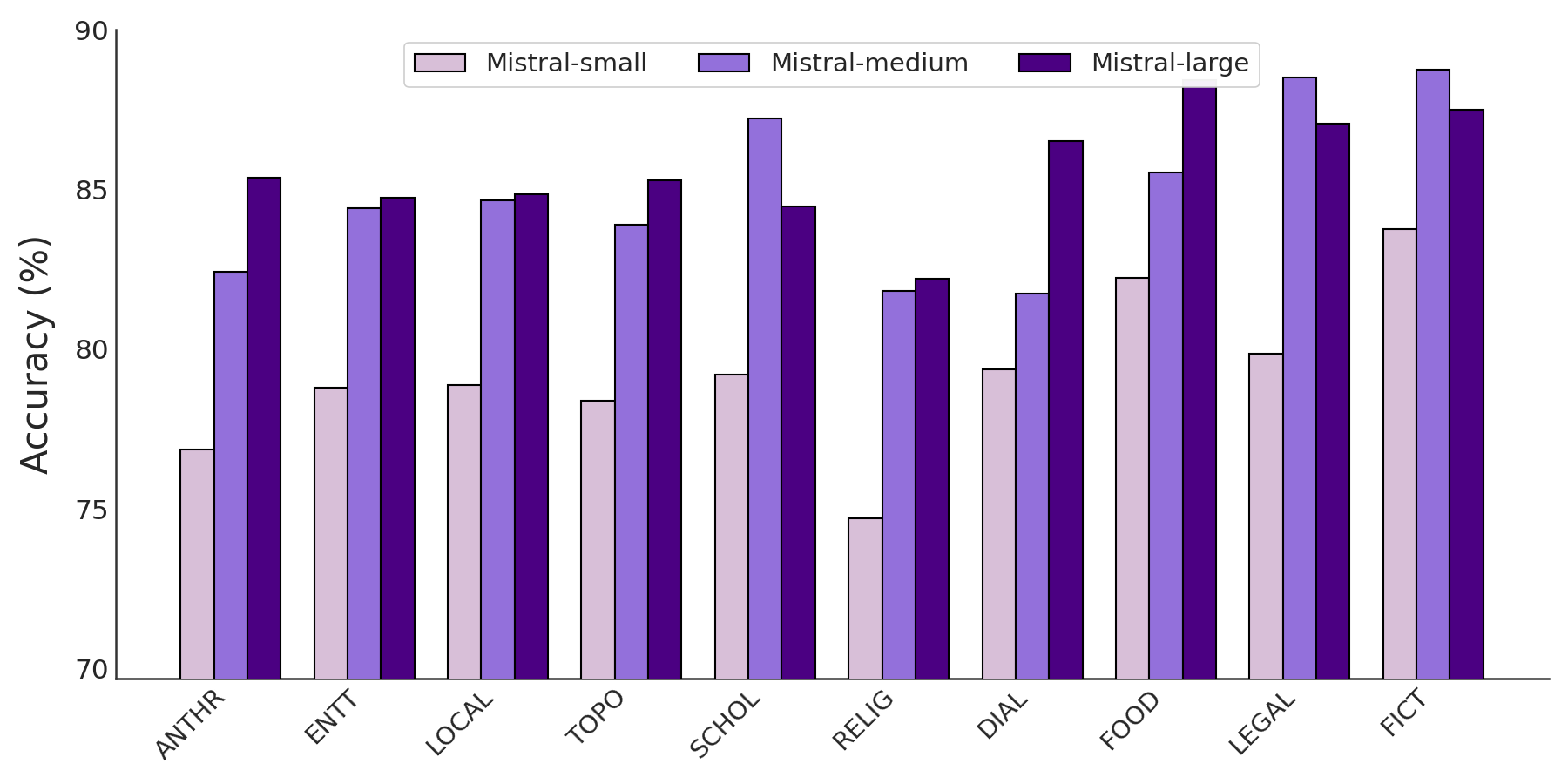}
    \caption{Performance of Mistral models on Latam Spanish data, with respect to the cultural element.}
    \label{fig:perf_per_cultural_element}
\end{figure*}

Figure \ref{fig:perf_per_cultural_element} shows the performances of the Mistral models for separated with respect to the cultural elements of the questions. 

\end{document}